\documentclass[letterpaper, 10 pt, conference]{ieeeconf}  

\IEEEoverridecommandlockouts                              

\usepackage[switch]{lineno}
\usepackage{graphics} 
\usepackage{epsfig} 
\usepackage{mathptmx} 
\usepackage{times} 
\usepackage{amsmath} 
\usepackage{amssymb}  
\usepackage{blindtext}
\usepackage{multicol}
\usepackage[bookmarks=true]{hyperref}
\usepackage{url}
\usepackage{subcaption}
\usepackage{bbm}
\usepackage{microtype}
\usepackage{graphicx}
\usepackage{booktabs}
\usepackage[utf8]{inputenc}
\usepackage[T1]{fontenc}
\usepackage[noadjust]{cite}
\usepackage{wrapfig,lipsum}
\usepackage{caption}
\usepackage{nicefrac}
\usepackage{microtype}
\usepackage{epstopdf}
\usepackage{color}
\usepackage{txfonts}
\usepackage[boxed]{algorithm2e}
\usepackage{pifont}
\usepackage{soul}

\newcommand{\cmark}{\ding{51}}%
\newcommand{\xmark}{\ding{55}}%
\newcommand\green[1]{{}{#1}} 
\newcommand\blue[1]{{}{#1}} 

\let\labelindent\relax
\usepackage[inline]{enumitem}

\usepackage{amsmath,amsfonts,bm}

\def\eqref#1{equation~\ref{#1}}

\def\1{\bm{1}}

\DeclareMathAlphabet{\mathsfit}{\encodingdefault}{\sfdefault}{m}{sl}
\SetMathAlphabet{\mathsfit}{bold}{\encodingdefault}{\sfdefault}{bx}{n}

\makeatletter
\DeclareRobustCommand\onedot{\futurelet\@let@token\@onedot}
\def\@onedot{\ifx\@let@token.\else.\null\fi\xspace}

\makeatother

\newcommand{\approptoinn}[2]{\mathrel{\vcenter{
  \offinterlineskip\halign{\hfil$##$\cr
    #1\propto\cr\noalign{\kern2pt}#1\sim\cr\noalign{\kern-2pt}}}}}

\title{\LARGE \bf
CLEVER: Stream-based Active Learning for Robust \\ Semantic Perception from Human Instructions}

\let\oldtwocolumn\twocolumn
\renewcommand\twocolumn[1][]{%
    \oldtwocolumn[{#1}{
    \begin{center}
           \includegraphics[width=\textwidth]{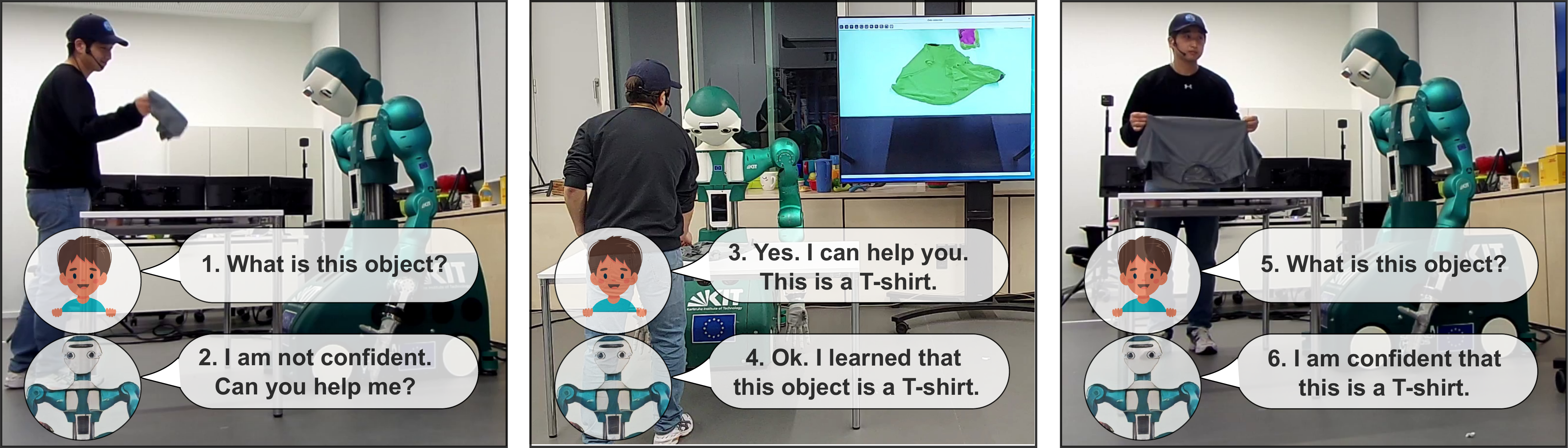}
           \captionof{figure}{\green{Imagine a robot encountering an unseen object, e.g., trained for apples but tested on a T-shirt. This problem of distribution shifts induce learning algorithms to typically fail. With the proposed system, CLEVER, the robot queries a human when the model is uncertain, and adapt itself to reduce that uncertainty. With human instructions, we demonstrate that such query-and-adaptation capabilities can improve the robustness of DNN-based semantic perception against distribution shifts.}}
           \label{fig1:teaser}
           \vspace*{-2mm}
    \end{center}
    }]
}

\author{Jongseok~Lee$^{1,2}$, Timo Birr$^{2}$, Rudolph Triebel$^{1,2}$ and Tamim Asfour$^{2}$ 
\thanks{$^{1}$Institute of Robotics and Mechatronics, German Aerospace Center (DLR), 82234 Weßling, Germany. $^{2}$Institute for Anthropomatics and Robotics, Karlsruhe Institute of Technology, 76131 Karlsruhe, Germany.}
\thanks{This work is supported by euROBIN (grant agreement 101070596), Inverse (grant agreement 101136067) and AiSac (Grant No. RS-2024-00441872).}
\thanks{Correspondence to {\tt\small jongseok.lee@dlr.de}}
}

\begin{document}


\maketitle
\thispagestyle{empty}
\pagestyle{empty}

\begin{abstract}


We propose CLEVER, an active learning system for robust semantic perception with Deep Neural Networks (DNNs). For data arriving in streams, our system seeks human support when encountering failures and adapts DNNs online based on human instructions. In this way, CLEVER can eventually accomplish the given semantic perception tasks. Our main contribution is the design of a system that meets several desiderata of realizing the aforementioned capabilities. The key enabler herein is our Bayesian formulation that encodes domain knowledge through priors. Empirically, we not only motivate CLEVER's design but further demonstrate its capabilities with a user validation study as well as experiments on humanoid and deformable objects. To our knowledge, we are the first to realize stream-based active learning on a real robot, providing evidence that the robustness of the DNN-based semantic perception can be improved in practice. The project website can be accessed at \url{https://sites.google.com/view/thecleversystem}.

\end{abstract}
\section{Introduction}

With deep neural networks (DNNs), the performance of computer vision increased dramatically, achieving impressive results in the semantic perception tasks, such as object classification, detection, and segmentation \cite{kirillov2023segment, oquab2024dinov}. However, such advancements in computer vision may not directly translate to the robotic semantic perception. This is because, in contrast to standard computer vision benchmarks, robots are situated in the physical world, where unpredictable events routinely occur and affect the robustness of the robot's understanding of its own environments. An example is distributional shift scenarios, where DNNs often make unexpected errors due to test conditions being underrepresented in the training data \cite{ancha2024deep, feng2023topology}. Thus, to achieve robust semantic perception, several probabilistic techniques have been investigated so far, so that with uncertainty estimates, robots can reason when to trust the predictions from DNNs and when not \cite{sirohi2023uncertainty, lee2022trust, ren2023robots, lakshminarayanan2017simple, gal2016dropout}.

\blue{In this paper, we build upon such probabilistic techniques and propose a system called CLEVER. The main idea behind CLEVER} is to not only obtain uncertainty estimates from DNNs for probabilistic predictions, but to further reduce the model's uncertainty by asking support from humans and adapting the model online. We achieve this query and adaptation through so-called stream-based active learning (AL) -- an autonomous learning paradigm that involves continuously selecting and labeling new data as they arrive in a stream, allowing for adaptation of the model online to changes in data distribution \cite{cacciarelli2024active}. The outcome of CLEVER is an adaptable DNN for semantic perception. \blue{For such capabilities, we equip CLEVER with a continuously adaptable DNN, a Bayesian learning algorithm, and an AL with temporal information. In particular, our Bayesian algorithm learns informative prior -- a probability distribution over the model parameters that incorporates domain knolwedge.}
\begin{table*}[ht!]
\centering
\caption{\blue{The proposed system addresses several desiderata for demonstrating a stream-based AL with real robots.}}
\begin{tabular}{ccccc}
\toprule
\multicolumn{1}{c}{} & \multicolumn{1}{c}{\textbf{\textbf{Both uncertainty }}} & \multicolumn{1}{c}{\textbf{Ability to ask help}} & \multicolumn{1}{c}{\textbf{Addresses catastrophic}} & \multicolumn{1}{c}{\textbf{Update DNNs fast,}}\\
\multicolumn{1}{c}{} & \multicolumn{1}{c}{\textbf{and generalization}} & \multicolumn{1}{c}{\textbf{and select samples}} & \multicolumn{1}{c}{\textbf{forgetting in DNNs}} & \multicolumn{1}{c}{\textbf{\textbf{e.g., less than 1 minute}}} \\
\midrule
\textbf{Continual learning} (i.e., \cite{blum2022self, frey2022continual})  & \xmark  & \xmark  & \cmark  & \cmark \\
\textbf{Existing stream-based AL}  (i.e., \cite{saran2023streaming, schmidt2023stream, narr2016stream})  & \xmark & \cmark  & \xmark  & \xmark \\
\textbf{Interactive learning} (i.e., \cite{azagra2020incremental, valipour2017incremental, schiebener2011segmentation})  & \xmark   & \xmark & \xmark  & \cmark \\
\textbf{Our system CLEVER}  & \cmark   & \cmark  & \cmark  & \cmark \\
\bottomrule
\end{tabular}
\label{table1:relatedworks}
\vspace*{-5mm}
\end{table*}

CLEVER meets several desiderata of stream-based AL with DNNs in practice. By learning priors, CLEVER is designed to generalize and estimate well-calibrated uncertainty, even with limited data availability. Such prediction capabilities are crucial to request support from humans ("query") only when necessary. CLEVER also addresses the issue of catastrophic forgetting, i.e., the tendency of DNNs to abruptly forget about previously learned tasks when continuously learning a new task \cite{wang2024comprehensive}. Furthermore, CLEVER can learn a new task in one minute by updating only the relevant parameters of DNNs online while using only fewer but most informative and diverse training data. In the experiments, we provide several ablation studies and comparative assessments to motivate our design choices. Finally, through a user validation study with 13 participants and the deployment of CLEVER on the humanoid robot ARMAR-6 \cite{asfour2019armar}, we demonstrate the enhanced robustness in semantic perception with robots.

\textbf{Contributions and major claims.} To the best of our knowledge, CLEVER is the first stream-based active learning system with DNNs, shown in a physical system for robotic perception tasks. Moreover, unlike existing works, we apply stream-based active learning for securing robustness in semantic perception tasks. To enable this novel capability, we identify new system requirements and challenges (Section \ref{sec:systemconcept}), followed by CLEVER's design that meets these requirements within a single framework (Section \ref{sec:method}). CLEVER is evaluated in response to these requirements. In particular, we show that our Bayesian formulation with learning-based priors enhances the practicality of CLEVER (Section \ref{sec:results1}). Through a user validation study that involves arbitrary objects (Section \ref{sec:results2}), and demonstrations on a humanoid robot for deformable object perception (Section \ref{sec:results3}), we create distributional shift scenarios for evaluation. Even under these challenging scenarios, we show that CLEVER can eventually accomplish the given perception tasks, improving the robustness of the DNN-based semantic perception in the real world.
\section{Related Work}

Our primary contribution is in the area of AL. For this, we bring Bayesian methods for neural networks and interactions with humans for robot learning. Thus, we locate our work within these areas. Tab.~\ref{table1:relatedworks} summarizes our main novelty.

\textbf{Stream-based active learning.} Active Learning (AL) is a paradigm in which a learning algorithm identifies the most useful unlabeled instances to learn from \cite{cacciarelli2024active}. In the literature, a pool of unlabeled instances is mostly assumed, resulting in the so-called pool-based AL \cite{Noseworthy21, feng2022bayesian, akcin23a}. In contrast, we focus on a setting in which data arrive in the stream, which is an underexplored area of AL \cite{cacciarelli2024active}. For robotic perception, the early attempts for stream-based AL relied on classical learning techniques such as Gaussian Processes \cite{triebel2016driven}, boosting \cite{mund2015active} and bagging \cite{narr2016stream}. Yet, the current de-facto standard in object recognition relies on deep learning, urging for extensions of stream-based AL to DNNs. Although current extensions \cite{saran2023streaming, schmidt2023stream} study the feasibility of stream-based AL using DNNs, the discussions therein are centered on strategies for informative data selection from the data stream. 

Indeed, the central objective of AL is to reduce the cost of labeling by querying and selecting the most informative data \cite{cacciarelli2024active}. In contrast, our focus is applying AL to  enhance the robustness of robotic perception by seeking human support and updating DNNs online. A similar use case was also previously mentioned by Triebel et al. \cite{triebel2016driven, narr2016stream}. However, due to a different focus, no real system was therein developed and evaluations were limited to showing sample efficiency, i.e., accuracy increase per newly added data points. Instead, we take a systems approach to the problem, thereby developing CLEVER that meets various requirements reported in Tab.~\ref{table1:relatedworks}.

\textbf{Bayesian adaptation of neural networks.} Learning semantics from new streams of observation is a crucial capability for our system. In robotics, such adaptations with DNN have previously been investigated \cite{blum2022self, frey2022continual}. Their findings suggest that continual learning during deployment improves the accuracy of the robot's perception when compared to fixed, pretrained DNNs. However, their applicability to stream-based AL is limited, since no uncertainty estimates are available for the query and selection step of AL. In contrast, our work explores Bayesian methods that are well suited for stream-based AL within a unified framework. For this, we extend our previous work \cite{schnaus2023learning} to stream-based AL. We  previously showed how continual learning can be performed while obtaining well-calibrated uncertainty estimates and generalization with DNNs using few data samples \cite{schnaus2023learning}. We point out that such properties are desired for developing a complete stream-based AL system with real robots.

\begin{figure*}[ht!]
    \centering
    \includegraphics[width=1.0\linewidth]{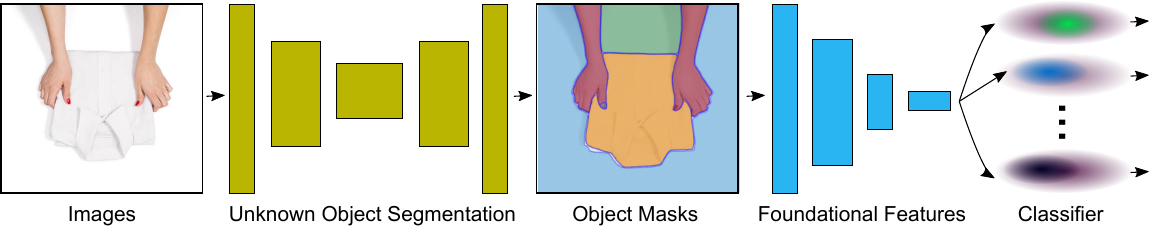}
    \caption{Our prediction model for stream-based AL using DNNs. For semantic segmentation, our pipeline combines an unknown object segmentation (Segment-Anything), foundational representations (DinoV2) and a classifier based on Bayesian Neural Networks (BNNs) with multilayer perceptron. BNNs are represented as Gaussian distribution over the model parameters.}
    \label{fig:fig2}
    \vspace*{-5mm}
\end{figure*}

\textbf{Robot learning from humans.} In robotics, many works on learning from demonstration focus on mapping the states of robots to actions \cite{ravichandar2020recent}. Yet, a recent work \cite{ren2023robots} shares a similar spirit to ours, i.e., an algorithm is designed to ask for human help using uncertainty estimates. However, their focus is on robot planning using language models. An idea on the correction of DNNs with language instructions from humans is also being explored for policy learning \cite{shi2024yell}. Many researchers have investigated incremental and interactive learning of new objects using human instructions \cite{azagra2020incremental, valipour2017incremental, schiebener2011segmentation}. Among them, we extend the works on a humanoid robot, ARMAR-III, where the robot demonstrated interactive learning of unknown objects from human instructions \cite{asfour2006armar, schiebener2011segmentation}. These early attempts showed that new rigid objects, such as books, can be learned using hand-crafted visual features and a k-nearest-neighbor classifier. Our extensions provide (a) the use of DNNs, (b) stream-based AL that reduces model uncertainty, and (c) fewer prior assumptions about the object. 

\section{System Concept and Challenges}
\label{sec:systemconcept}

Whenever uncertain, our system seeks human support and improves itself online based on human instructions. The main problem encountered is distribution shifts that the robot inevitably encounters during the deployment of a DNN-based model, which ultimately produces misleading and overconfident predictions. For example, a robot may face unknown objects. Imagine a robot trained to classify apples but spots bananas during deployment. Deformable objects present similar challenges if the induced deformation in the object's shape is underrepresented in the training data. Given this problem, the concept of our system is to ask for help from humans and adapt the model online (see Fig. \ref{fig1:teaser}).  

\blue{There are several challenges (see Tab. \ref{table1:relatedworks}). First, the system should know when the model is uncertain. This requires a probabilistic treatment that provides well-calibrated uncertainty estimates under distribution shifts. Given a training data $\mathcal{D}$ and a DNN, $f_{\boldsymbol{\theta}}$, where all learnable parameters are stacked as a vector $\boldsymbol{\theta}$, probabilistic treatments of the given problem infer the posterior $p(\boldsymbol{\theta} | \mathcal{D})$ and compute a predictive distribution $p(\mathbf{y}^* | \mathbf{x}^*, \mathcal{D})$ for new test datum $(\mathbf{x}^*, \mathbf{y}^*) \notin \mathcal{D}$. Here, acquired data cannot be of large amounts as we rely on manual human instructions. Thus, we need DNNs that generalize with small amount of data. Second, the system should adequately support queries from the human and a decision-making framework that enables effective instruction of the robot by a human. The former minimizes the need for repeated human querying. We select a subset of data $\overline{\mathcal{D}}^* \subseteq \mathcal{D}^*$ that is the most informative to learn. This reduces the training time. Third, the model should continually learn and produce accurate predictions without forgetting. Lastly, for demonstrations on a humanoid, DNNs are to be trained fast, e.g., under one minute. To achieve this, CLEVER trains only a subset of model parameters $\overline{\boldsymbol{\theta}} \subseteq \boldsymbol{\theta}$ and selected data $\overline{\mathcal{D}}^*$. }


\begin{figure*}[ht!]
    \centering
    \includegraphics[width=1.0\linewidth]{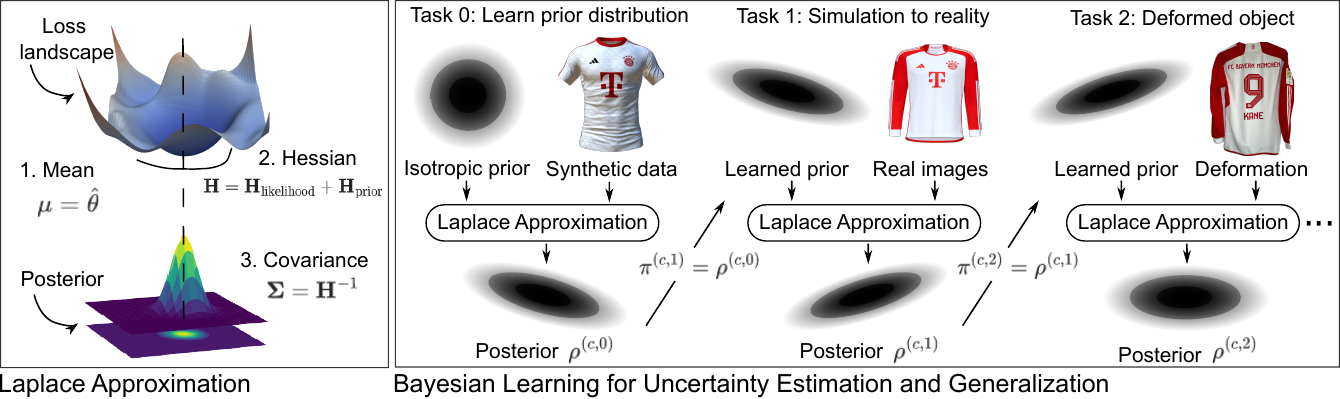}
    \caption{The proposed pipeline to learn a prediction model for stream-based AL. Left: Laplace Approximation (LA) infers the posterior distribution of a DNN as a Gaussian distribution. Right: Using LA to obtain BNNs and further exploiting training data that encodes our domain knowledge, we learn an informative prior from a posterior of a previous task. \green{Bayesian learning in sequence addresses the potential prior misalignment where humans provide the relevant task and data to learn the prior.}}
    \label{fig:fig3}
    \vspace*{-5mm}
\end{figure*}

\section{CLEVER -- A Stream-based Active Learner}
\label{sec:method}

Having discussed the system concept and challenges, we describe the design of CLEVER in more detail.

\subsection{Underlying prediction model for continual adaptation}

Our first component is a prediction model for semantic perception (see Fig.~\ref{fig:fig2}). The overall model takes images as input and outputs segmentation masks with their associated output labels. For this, we rely on an unknown object segmentation such as Segment-Anything \cite{kirillov2023segment} to generate the segmentation masks. A tracker can be combined to achieve a faster runtime of the model \cite{kirillov2023segment}. Then, we stack the obtained masks and extract features using foundational models such as DinoV2 \cite{oquab2024dinov}. The obtained features are used as input of a classifier, which generates output labels for each segmentation mask. In contrast to existing models such as Mask-RCNN, our construction allows open-set extraction of segmentation masks and state-of-the-art visual features using pre-trained models. This means we only need to train a classifier. In lieu of sophisticated pipelines for semantic segmentation such as Mask-RCNN, we focus on AL for classification tasks while also obtaining detection and segmentation results. 

Next, we present the design of our classifier for stream-based AL. The classifier learns one multilayer perceptron (MLP) per object class. We call each of these MLPs the \blue{head}s of our classifier. If we had an apple and a banana, our classifier would have two \blue{head}s, each responsible for only one object class. In this way, a multi-class classification task is tackled using a combination of binary classifiers with calibrated uncertainties. The proposed construction brings two advantages. First, we can mitigate catastrophic forgetting by design. As we create a new \blue{head} for a new incoming class, learning new objects does not affect the previously learned \blue{head}s. Moreover, in each learning cycle, we can update only a \blue{head} of our classifier. For example, if a \blue{head} responsible for an apple exhibits high uncertainty, we can only learn to classify an apple better. Once the classifier is confronted with an unknown object, we can add and learn one new \blue{head}. Training a smaller model can be more efficient. \blue{Achieving these results with a multi-class classifier might be difficult.}

Within this construction, we apply probabilistic inference on all MLPs to obtain Bayesian Neural Networks (BNNs), which can provide well-calibrated uncertainty under distribution shifts. \blue{Let us define the training data for the classifier as $\mathcal{D}=\left((\mathbf{x}_i, \mathbf{y}_i)\right)_{i=1}^N$ where the inputs $\mathbf{x}_i$ are the extracted features.} We use superscripts $(c, t)$ to denote the available C classes and T tasks, respectively. Then, a classifier $f_{\boldsymbol{\theta}}$ is now divided into several binary classifiers $f_{\boldsymbol{\theta}^{(c, t)}}$ that output $\mathbf{y}_c \in \left\{ 0, 1 \right\}$. The corresponding DNN's learnable parameters $\boldsymbol{\theta}^{(c, t)}$ belong to the \blue{head} c for the task t, which is obtained using the relevant training data $\mathcal{D}^{(c, t)}$. BNNs apply probabilistic inference to DNNs, e.g., models are represented by the posteriors $\rho^{(c, t)}=p(\boldsymbol{\theta}^{(c, t)} | \mathcal{D}^{(c, t)})$. BNNs predict through marginalization:
\begin{equation}
\label{eq:1}
p(\mathbf{y}^*_c=1|\mathbf{x}^*, \mathcal{D}^{(c, t)}) = \int p(\mathbf{y}^*_c|\mathbf{x}^*,\boldsymbol{\theta}^{(c, t)})p(\boldsymbol{\theta}^{(c, t)}|\mathcal{D}^{(c, t)}) d\boldsymbol{\theta}^{(c, t)}.
\end{equation}
Intuitively, instead of one model that best fit the data, BNNs pay tribute to the model uncertainty for predictions. The probabilities obtained better reflect the true belief in the class $\mathbf{y}^*_c$ than the often used softmax scores that tend to bias towards higher probabilities, or overconfident predictions \cite{lee2020estimating, schnaus2023learning}.

We then combine the calibrated binary classifiers using BNNs for handling the given multi-class classification task. Given a single test input $\mathbf{x}^*$, we obtain a vector of probabilities from all \blue{head}s: $\boldsymbol{p} = (p_1, p_2, ..., p_C)$ where $p_c = p(\mathbf{y}^*_c=1|\mathbf{x}^*, \mathcal{D}^{(c, t)})$. Then, we pick the predicted class label by:
\begin{equation}
\label{eq:2}
    \mathbf{y}^* = \text{argmax}_c(\boldsymbol{p}) \ \text{with} \ p(\mathbf{y}^*=c|\mathbf{x}^*) = p_{c=\mathbf{y}^*}.
\end{equation}
In this way, we select one output of a \blue{head} with the highest probabilities and choose the corresponding class label and probability score. We do not consider one vs. rest multi-class strategy. Hence, the vector $\boldsymbol{p}$ itself is not a valid probability distribution. For querying from humans and selecting the most informative data using uncertainty estimates, a valid probability distribution for the most likely class is sufficient. Thus, we do not require probabilities for all classes at once. Moreover, the given design choice enables our system to be more efficient when training, because we do not need to update all the \blue{head}s for training a single \blue{head} -- an assumption in one vs. rest multi-class strategy. Instead, we can update each \blue{head} individually for efficiency.

\subsection{Bayesian learning for uncertainty and generalization}

We now present our training pipeline for the aforementioned probabilistic model. Our goal is to obtain the posteriors $\rho^{(c, t)}$ for the estimation of the uncertainty. We also aim to address the challenge of generalization under a small data regime with DNNs. For this, our main idea is to learn the prior distribution over the parameters of DNNs. To explain, according to Bayes rule, the posteriors are proportional to the likelihood $\prod_{i=1}^N p(\mathbf{y}_i | \mathbf{x}_i, \boldsymbol{\theta}^{(c, t)})$ and the prior $\pi^{(c, t)} = p(\boldsymbol{\theta}^{(c, t)})$, i.e., $\rho^{(c, t)} = p(\boldsymbol{\theta}^{(c, t)} | \mathcal{D}^{(c, t)}) \propto \pi^{(c, t)} \prod_{i=1}^N p(\mathbf{y}_i | \mathbf{x}_i, \boldsymbol{\theta}^{(c, t)})$. Due to the non-linearity of DNNs, no closed-form solution exists for the posteriors. Thus, we need approximate Bayesian inference -- a set of algorithms that approximate the intractable posteriors. For this, the first step is to define the priors over the DNN parameters. Traditionally, a zero-mean isotropic Gaussian prior -- an uninformative prior that regularizes the overall model -- was seen to be sufficient for DNNs. However, when no large amounts of data are available, the likelihood no longer dominates the posterior. Thus, specifying an informative prior can improve the approximate Bayesian inference \cite{schnaus2023learning}. Fig. \ref{fig:fig3} shows our pipeline, which we later combine with a generalization framework called the PAC-Bayes theory.

The first step of our pipeline is to learn an initial prior distribution offline using synthetic data (task 0 in Fig. \ref{fig:fig3}). Synthetic data for object recognition can be generated by either photorealistic synthesizers such as BlenderProc2 or generative models such as StableDiffusion with relevant prompts like "A jersey on a table". Synthetic data has the advantage that large amounts of annotated training data can be generated in a cost-effective manner. Hence, for all the known classes of objects, we generate training data $\mathcal{D}^{(c, t=0)}$. Now, in order to learn an initial prior, we apply Laplace Approximation (LA). LA is an approximation inference method that imposes Gaussian distribution on the DNN parameters around a local mode \cite{lee2020estimating}. The obtained posterior has its mean as the maximum-a-posterior (MAP) estimates $\boldsymbol{\mu}^{(c, t)} = \hat{\boldsymbol{\theta}}^{(c, t)}$, which can be obtained using the standard DNN training procedure with a cross-entropy loss. In LA, the covariance of the posterior $\mathbf{\Sigma}^{(c, t)}$ is estimated by an inverse of a loss landscape's Hessian $\mathbf{H}^{(c, t)}$, i.e., a second order derivative of log posterior w.r.t the DNN parameters $\boldsymbol{\theta}^{(c, t)}$. By definition, $\mathbf{H}^{(c, t)} = \mathbf{H}_{\text{likelihood}}^{(c, t)} + \mathbf{H}_{\text{prior}}^{(c, t)}$. Assuming an isotropic prior,
\begin{flalign}
\label{eq:3}
    &\text{Prior:} \quad \pi^{(c, 0)} = \mathcal{N}(\boldsymbol{0}, \gamma \mathbf{I}), &&\\
    &\text{Posterior:} \quad \rho^{(c, 0)} \approx \mathcal{N}(\boldsymbol{\hat{\theta}}^{(c, 0)} , (\mathbf{H}_{\text{likelihood}}^{(c, 0)} + \gamma \mathbf{I})^{-1}), &&\nonumber
\end{flalign}
are the prior-and-posterior pairs. The posteriors at $t=0$ is then used as priors for task $t=1$. \green{We use a layer-wise Kronecker factorization \cite{schnaus2023learning} to compute the Hessian or the covariance.}

Having learned the priors, we now iterate the learning process online.  Examples of incoming tasks are semantic segmentation on real images, or recognition of deformable objects, as depicted Fig. \ref{fig:fig3}. In each step, the approximated posteriors from the previous tasks are used as the priors for the current learning tasks. Intuitively, as we keep learning one object class per \blue{head}, such Bayesian learning results in positive transfers across the tasks, i.e., posteriors that classify folded clothes helps in learning unfolded clothes, even with small amounts of data. For this, we repeat LA and obtain:
\begin{flalign}
\label{eq:4}
    &\text{Prior:} \quad \pi^{(c, t)}  = \mathcal{N}(\boldsymbol{\hat{\theta}}^{(c, t-1)}, (\mathbf{H}^{(c, t-1)})^{-1}), && \\
    &\text{Posterior:} \quad \rho^{(c, t)}  \approx \mathcal{N}(\boldsymbol{\hat{\theta}}^{(c, t)}, (\mathbf{H}_{\text{likelihood}}^{(c, t)} + \mathbf{H}^{(c, t-1)})^{-1}). &&\nonumber
\end{flalign}
The prior-and-posterior pairs are obtained with small modifications to LA. First, the maximum-a-posterior estimates are computed using a more expressive prior, instead of initializing around zero with one variance for all DNN parameters. Second, for approximating the posteriors, the Hessian from the posterior of previous task is used instead of an isotropic term $\gamma \mathbf{I}$. \blue{Our pipeline adapts our previous method \cite{schnaus2023learning} to better fit our use case. Unlike previously, we do not attempt to transfer across \blue{head}s. This enables class-independent learning. Moreover, we embed the domain knowledge with synthetic data at task 0, instead of foundational models.}

For these steps of Bayesian learning, we can explicitly design for improving generalization of the model under small data regime. We achieve this by introducing hyperparameters $\tau, \alpha$ and $\beta$ s.t. $\mathbf{H} = \tau (\beta \mathbf{H}_{\text{likelihood}} + \alpha \mathbf{H}_{\text{prior}})$, and optimizing for a generalization objective called PAC-Bayes bounds. Note that we dropped the superscript $(c,t)$ for notation simplicity when explaining PAC-Bayes theory. The hyperparameters $\alpha$ and $\beta$ decide how much weight should be given to the previous task (prior) and the current data at hand (likelihood). If more weight is given to the current data, the model may overfit, while more weight to the prior may result in underfitting. The tempering term $\tau$ scales the overall posterior. \blue{We pick these hyperparameters by minimizing an upper bound to the expected loss $\mathbb{E}_{\boldsymbol{\theta} \sim \rho}[\mathcal{L}^{l}_{P}(f_{\boldsymbol{\theta}})]$ on the true distribution $P$. A true expected loss incurs over the true data distribution $P$ -- not only on training and test data. Such generalization bounds depend on empirical loss on the training data $\mathbb{E}_{\boldsymbol{\theta} \sim \rho}[\hat{\mathcal{L}}^{l}_{\mathcal{D}}(f_{\boldsymbol{\theta}})] =\frac{1}{N}\sum_{i = 1}^N \hat{\mathcal{L}}^{l}_{\mathcal{D}}(f_{\boldsymbol{\theta}})$ and the KL-divergence of the priors and the posteriors. Our KL-divergence is dependent on $\tau$, $\alpha$ and $\beta$. For $\epsilon > 0$, the so-called PAC-Bayes bounds are defined as:} 
\begin{align}
    \label{eq:5}
    P_{\mathcal{D} \sim P} &( \forall \rho \ll \pi:\ \mathbb{E}_{\boldsymbol{\theta} \sim \rho}[\mathcal{L}^{l}_{P}(f_{\boldsymbol{\theta}})] \\
    &\leq \delta(\mathbb{E}_{\boldsymbol{\theta} \sim \rho}[\hat{\mathcal{L}}^{l}_{\mathcal{D}}(f_{\boldsymbol{\theta}})], \text{KL}(\rho\|\pi), N, \varepsilon) ) \geq 1 - \varepsilon. \nonumber
\end{align}
\blue{For more details, we refer to our previous work \cite{schnaus2023learning} where we devised a computationally tractable method for LA.}

\subsection{A temporal active learning system with humans}

\setlength{\textfloatsep}{0pt}
\begin{algorithm}
   \caption{CLEVER - stream-based active learner.}
   \label{alg:1}
   \SetAlgoLined
   \BlankLine
   \Begin{
   \tcp{Initialization}
   $\rho^{(c, 0)}$ $\leftarrow$ prior($\mathcal{D}^{(c,0)}, \pi^{(c,0)}$) $\forall c$ \tcp*{\scriptsize{Eq. \ref{eq:3}}}
   
   $\alpha^{(c, 0)}, \beta^{(c, 0)}, \tau^{(c, 0)}$ $\leftarrow$ pac-bayes($\bullet$) $\forall c$ \tcp*{\scriptsize{Eq. \ref{eq:5}}}
   \BlankLine
   \tcp{Main Loop}
   \While(){\emph{incoming image stream}}
   {
       $p_c | \mathbf{x}^*_{k}$ $\leftarrow$ marginalization($\mathbf{x}^*_{k}$) $\forall c$ \tcp*{\scriptsize{Eq. \ref{eq:1}}}
       
       $p_c | \mathbf{x}^*_{1:k-1}$  $\leftarrow$ filter($p_c|\mathbf{x}^*_{k}$) $\forall c$ \tcp*{\scriptsize{Eq. \ref{eq:6} (option).}}

       $\mathbf{y}^*, p_{c=\mathbf{y}^*}$ $\leftarrow$ prediction($\boldsymbol{p} | \mathbf{x}^*_{1:k-1}$) \tcp*{\scriptsize{Eq. \ref{eq:2}}}

       \If(){\emph{query humans for \blue{head} c}}
       {
      $\mathcal{D}^{(c, t)}_{new}$ $\leftarrow$ human-instruction() \tcp*{\scriptsize{Fig. \ref{fig1:teaser}}}
      
      $\mathcal{\bar{D}}^{(c,t)}$ $\leftarrow$ selection($\mathcal{D}^{(c, t)}_{new}$) \tcp*{\scriptsize{Eq. \ref{eq:7}}}
       
      $\rho^{(c, t)}$ $\leftarrow$ posterior($\mathcal{\bar{D}}^{(c,t)}, \pi^{(c,t)}$) \tcp*{\scriptsize{Eq. \ref{eq:4}}}

      $\alpha^{(c, t)}, \beta^{(c, t)}, \tau^{(c, t)}$ $\leftarrow$ pac-bayes($\bullet$) \tcp*{\scriptsize{Eq. \ref{eq:5}}}
       }
   }
   }
\end{algorithm}

\begin{figure*} 
  \begin{subfigure}[b]{0.25\linewidth}
    \centering
    \includegraphics[width=1\linewidth]{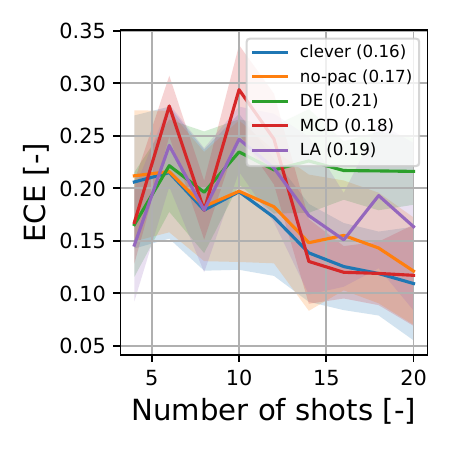} 
    \caption{Test ECE} 
    \label{fig4:a} 
  \end{subfigure}
  \begin{subfigure}[b]{0.245\linewidth}
    \centering
    \includegraphics[width=1\linewidth]{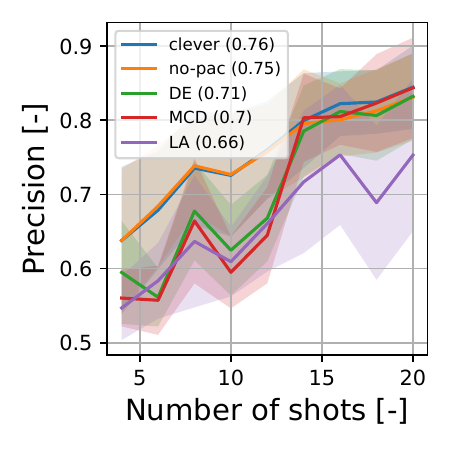} 
    \caption{Test Accuracy} 
    \label{fig4:b} 
  \end{subfigure} 
  \begin{subfigure}[b]{0.245\linewidth}
    \centering
    \includegraphics[width=1\linewidth]{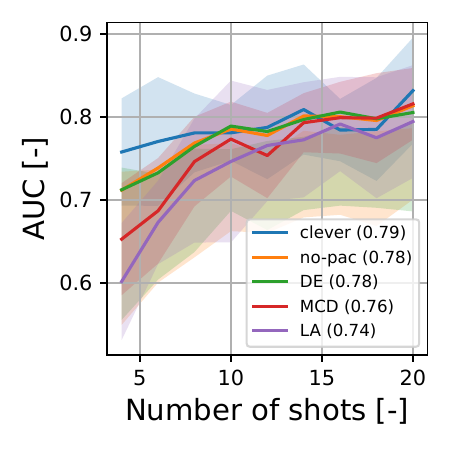} 
    \caption{Test AUC} 
    \label{fig4:c} 
  \end{subfigure} 
  \begin{subfigure}[b]{0.245\linewidth}
    \centering
    \includegraphics[width=1\linewidth]{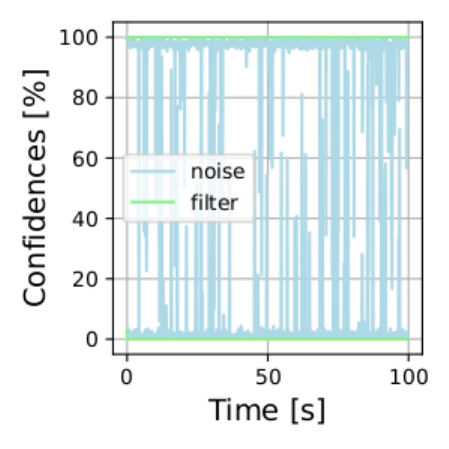} 
    \caption{Temporal Combination} 
    \label{fig4:d} 
  \end{subfigure}
  \begin{subfigure}[b]{0.245\linewidth}
    \centering
    \includegraphics[width=1\linewidth]{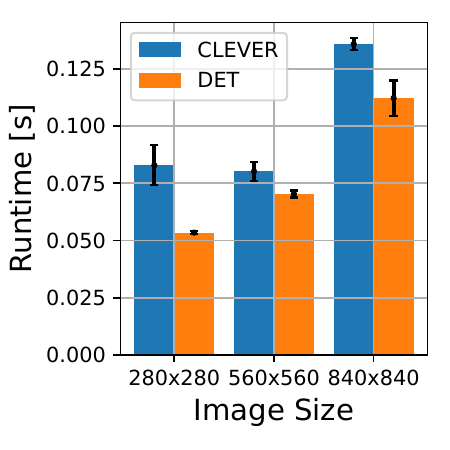} 
    \caption{Run-time Analysis} 
    \label{fig4:e} 
  \end{subfigure} 
  \begin{subfigure}[b]{0.245\linewidth}
    \centering
    \includegraphics[width=1\linewidth]{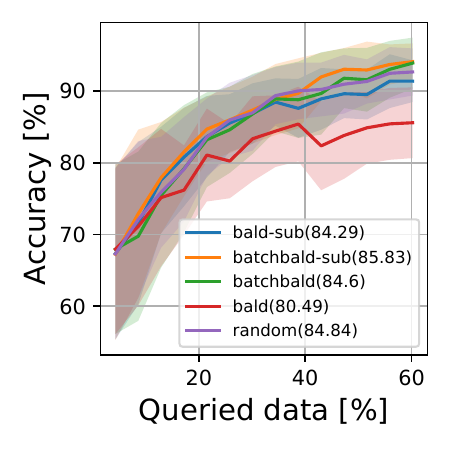} 
    \caption{Acquisition Functions} 
    \label{fig4:f} 
  \end{subfigure}
  \begin{subfigure}[b]{0.245\linewidth}
    \centering
    \includegraphics[width=1\linewidth]{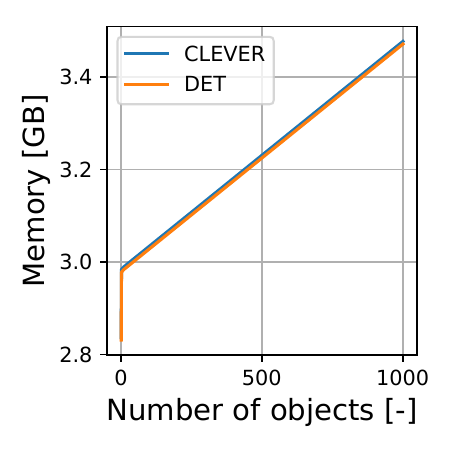} 
    \caption{Memory Analysis} 
    \label{fig4:g} 
  \end{subfigure} 
  \begin{subfigure}[b]{0.245\linewidth}
    \centering
    \includegraphics[width=1\linewidth]{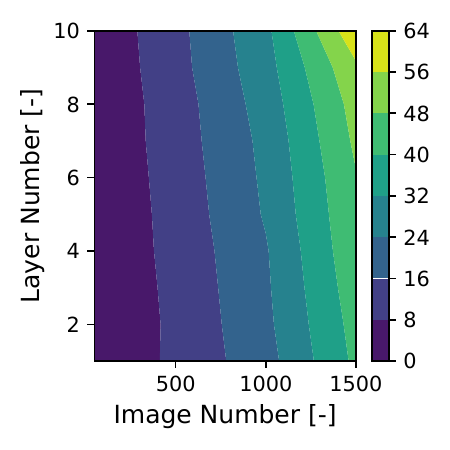} 
    \caption{Training Analysis} 
    \label{fig4:h} 
  \end{subfigure}
  \caption{Results from ablation studies and comparative assessments of various design choices. \blue{The unit for (h) is seconds.}}
  \label{fig4}
  \vspace*{-6mm}
\end{figure*}

The remaining challenges are to develop a system that (a) asks for help from humans and (b) selects the most informative samples to learn from. For both components, we combine the temporal information inherent in data streams. 

Our query strategy involves a recursive rule that keeps the current probabilities about an object class given all measurements until step k: $p(\mathbf{y}^*_{c}=1|\mathbf{x}^*_{1:k})$. Defining $l(\bullet) = \text{log}p(\bullet)[1-p(\bullet)]^{-1}$ from which we can retrieve the current probabilities $p(\bullet) = 1 - (1+\text{exp}[l(\bullet)])^{-1}$, our recursive form based on a binary state Bayes filter is given by:
\begin{equation}
\label{eq:6}
    l(\mathbf{y}^*_{c}=1|\mathbf{x}^*_{1:k}) = l(\mathbf{y}^*_{c}=1|\mathbf{x}^*_k) + l(\mathbf{y}^*_{c}=1|\mathbf{x}^*_{1:k-1})-l(\mathbf{y}^*_{c}=1).
\end{equation}
The obtained probabilities are converted to a normalized entropy as a measure of uncertainty, and we use a pre-defined threshold for query decisions \cite{triebel2016driven}. Here, temporal information may filter the outliers and augment the tracking of object segmentation. Our design choice on utilizing binary classifiers per \blue{head} also enables us to simplify the incorporation of temporal information. Instead of complex models such as Dirichlet, we only modified the algorithm behind the well-known probabilistic grid map \cite{burgard1996estimating} that tracks uncertainty on binary states such as occupied or non-occupied space. 

Next, given a human demonstration, AL selects the most informative data. For CLEVER, such a selection results in smaller training data, which makes the continual adaptation of the underlying model more efficient. We utilize the so-called BatchBald \cite{kirsch2019batchbald} to select the top $\bar{B} \subset B^{(c,t)}$ data points:
\begin{multline}
\label{eq:7}
    \mathcal{A}_{\text{batchbald}}(\mathbf{x}^*_{1:B}, \rho^{(c,t)}) = \mathbb{I}(\mathbf{y}^*_{c,1:B}, \boldsymbol{\theta}^{(c,t)} | \mathbf{x}^*_{1:B}, \mathcal{D}^{(c,t)}_{\text{new}}) \\
    = \mathbb{H}(\mathbf{y}^*_{c,1:B}|\mathbf{x}^*_{1:B}, \mathcal{D}^{(c,t)}_{\text{new}}) - \mathbb{E}_{\rho^{(c,t)}}\mathbb{H}(\mathbf{y}^*_{c,1:B}|\mathbf{x}^*_{1:B}, \boldsymbol{\theta}^{(c,t)}, \mathcal{D}^{(c,t)}_{\text{new}}).
\end{multline}
Intuitively, this acquisition function examines the mutual information $\mathbb{I}$ between the multiple model predictions and the model parameters. Such mutual information is obtained by approximations to the entropy terms $\mathbb{H}$. The coupling between the model predictions for a batch of data points and the model parameters is captured, and data points with high mutual information would inform us more about the true model parameters. This allows us to again combine temporal information by considering a batch of data points. We further combine a subsampling strategy \cite{feng2022bayesian}, that is, we randomly select a subset from the demonstration data before applying BatchBald. This ensures the diversity of the samples while speeding up the computations of BatchBald.

\subsection{\green{System overview and implementation details}}
\green{Alg. \ref{alg:1} provides an overview of CLEVER, which shows how all the components are integrated into a single algorithm.} For each task, humans instruct the semantic information about an object using speech and demonstrate the given object from different viewpoints. Negative examples from the previous demonstrations are also provided for training each \blue{head}. For unknown objects, we add a new \blue{head} and start the learning process with an isotropic prior. In Alg. \ref{alg:1}, $\bullet$ are used to indicate any arbitrary but relevant inputs to a function.
\section{Results}
\label{sec:results}

\subsection{Ablation studies and comparative assessments}
\label{sec:results1}
We provide ablation studies and comparative assessments to provide insights into the design of CLEVER. The analysis is focused on the continual learning architecture, the impact of using posteriors as informative priors, and combining temporal information. In particular, we focus on how our design choices address the challenges listed in Tab. \ref{table1:relatedworks}. For this, we collect a dataset from ARMAR-6, which consists of images from ten household objects. CAD models of these objects are obtained using an Android app called MagiScan. We also used StableDiffusion for synthetic data generation. For each object, we provided nine sequences of human demonstrations with 250 images each. We varied the difficulty level by providing more deformations to the objects, for example. We note that the dataset will be released to the public. There were no open-source datasets which fit our stream-based AL scenario due to the human element. We assumed that segmentation is given by foundational models like SAM, and mainly evaluate the underlying classifier. 

\begin{figure*}
    \centering
    \includegraphics[width=1\linewidth]{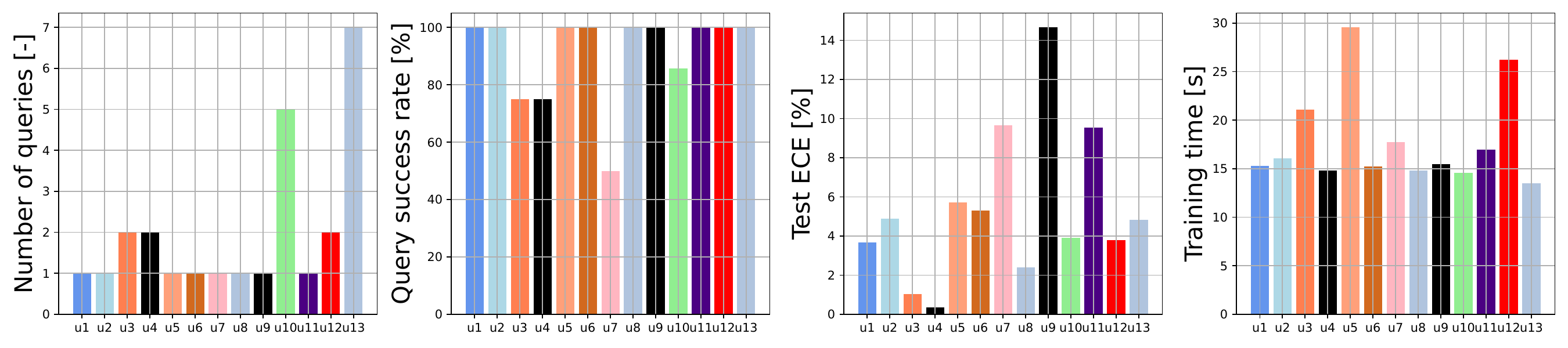}
    \caption{Results of a complete open-set evaluation with 13 users (x-axis). Number of queries to semantically segment the objects with more than 85$\%$ confidence (lower the better), query success rate (higher the better), ECE as a measure of uncertainty (lower the better), and training time (lower the better) are reported from the experiments per user. These experiments validate that various users can perform active learning with CLEVER for semantic segmentation under open-set conditions.}
    \label{fig:5}
    \vspace*{-4mm}
\end{figure*}

\blue{\textbf{Continuously adaptable model.}} First, we present our analysis on CLEVER's ability to perform continual learning. We claimed that the proposed design enables training of DNNs in less than one minute, while addressing the catastrophic forgetting problem. To evaluate the training time, we vary the number of layers from one to ten, and also vary the number of training data points up to 1500. Because we only train the MLP while fixing the representations from a foundational model, the results show that our classifiers can be updated in less than 1 minute (Fig. \ref{fig4:h}). We note that a three-layered MLP is used for all other experiments. Regarding the catastrophic forgetting, CLEVER adapts a progressive architecture where new \blue{head}s are trained for new incoming object categories. \green{By design, such architecture-based approaches mitigate the forgetting. However, growing the DNN architecture may hurt the computational scalability. To evaluate such computational scalability, we grow the DNN architecture to accommodate 1000 object categories. Comparisons in terms of memory and runtime are provided with CLEVER without probabilistic treatments (denoted DET). The results are depicted in Fig. \ref{fig4:g} and \ref{fig4:e}. Without an elaborated mechanism to mitigate catastrophic forgetting, CLEVER can learn 1000 object categories with less than 4GB of GPU peak memory and interactive frame rates. We also note that forgetting mechanisms can be introduced when an application scenario demands many more object categories.}

\blue{\textbf{Bayesian learning algorithm.}} Secondly, we examine the influence of prior learning method for both uncertainty and generalization under small data regime. For this, we split the dataset into training and test set with a ratio of 8:2. Then, we randomly pick N-shot images per object category up to 20 images. Using Google’s uncertainty baseline, we evaluate various models (MC Dropout \cite{gal2016dropout} as MCD, Deep Ensemble as DE \cite{lakshminarayanan2017simple}, Laplace Approximation as LA \cite{lee2020estimating}, and no-pac means CLEVER without PAC-Bayes optimization \cite{schnaus2023learning}) with expected calibration error (ECE), \blue{precision} and area under ROC curve (AUC) as the standard evaluation metrics. The findings are shown in Fig. \ref{fig4:c}, \ref{fig4:b} and \ref{fig4:c}, where the chosen metrics are reported by averaging over the object categories. Five random seeds were used. The results show that CLEVER can outperform the chosen baselines. In particular, comparisons to LA and no-pac show that learning the prior from simulation, and optimizing for a generalization bound can improve the performance for the chosen evaluation scenario of stream-based AL. 

\blue{\textbf{Temporal information.}} Third, we examine the idea of combining the temporal information when selecting the subset of images to learn more efficiently, and also deciding to query. For the former, we choose a uniform sampling, BALD, BatchBALD and their combinations with sub-sampling as baseline acquisition functions. For the latter, we train a model to provide noisy confidence estimates, and display filtered output against the raw output from a single query step of 100s. The same train-test split of 8:2 was used for the comparisons on acquisition functions. A total of 15 query steps were generated for all object categories. Five random seeds were used to obtain the results. For the combination of temporal data via filtering, we observe that noises can be removed (Fig. \ref{fig4:d}). Moreover, BatchBALD with the sub-sampling strategy, as the acquisition function looks at the batch of data points to measure information gain, outperformed other baselines in Fig. \ref{fig4:f}. These findings motivate our design choice of integrating temporal information for performance improvements.

\begin{table}
	\setlength{\tabcolsep}{0.5em}
	\centering
	\caption{Query success rates, ECE, accuracy, and number of queries to reach 85$\%$ accuracy are reported.}
	\label{table:final:performance}
	\scalebox{1.0}{
		\begin{tabular}[ht]{ccccc}
			\toprule
			  & \textbf{Success rate} & \textbf{ECE} & \textbf{Precision} & \textbf{Nr. Queries} \\
			\midrule
			CLEVERv1 & \textbf{0.887$\pm$0.049} & \textbf{0.060$\pm$0.017} & \textbf{0.933$\pm$0.020} & \textbf{1.539$\pm$0.544}\\ 
            CLEVERv2 & {0.826$\pm$0.053} & {0.092$\pm$0.036} & {0.900$\pm$0.034} & {2.440$\pm$0.866}\\ 
			Vanilla & 0.801$\pm$0.054 & 0.177$\pm$0.021 & 0.817$\pm$0.039 & 4.320$\pm$0.992\\
			\bottomrule
		\end{tabular}
	}
\end{table}%

\blue{\textbf{Evaluation.}} Finally, we examine the final performance (see Tab. \ref{table:final:performance}). We assume that images arrive in a batch of streams over nine subsequent demonstrations with increasing levels of difficulty. \blue{Metrics of choice were query success rate, i.e., if the model queried correctly, average ECE and precision, and the number of queries required to reach more than 85 $\%$ precision. These metrics capture several requirements of a stream-based AL system.} 40 data points were selected for training in each demonstration out of 250 data points so that the training terminates in less than a minute. Five random seeds were used. We compared three baselines. Vanilla corresponds to a deterministic model of CLEVER without any priors. Version 1 used the full formulation of the prior with both mean and covariance, while version 2 only utilized the mean by pretraining with the given synthetic data. We observe the gradual increase in performance in all metrics. These results justify our design choices, in particular, the prior learning with the targeted synthetic data for stream-based AL.


\begin{figure}
    \centering
    \includegraphics[width=1\linewidth]{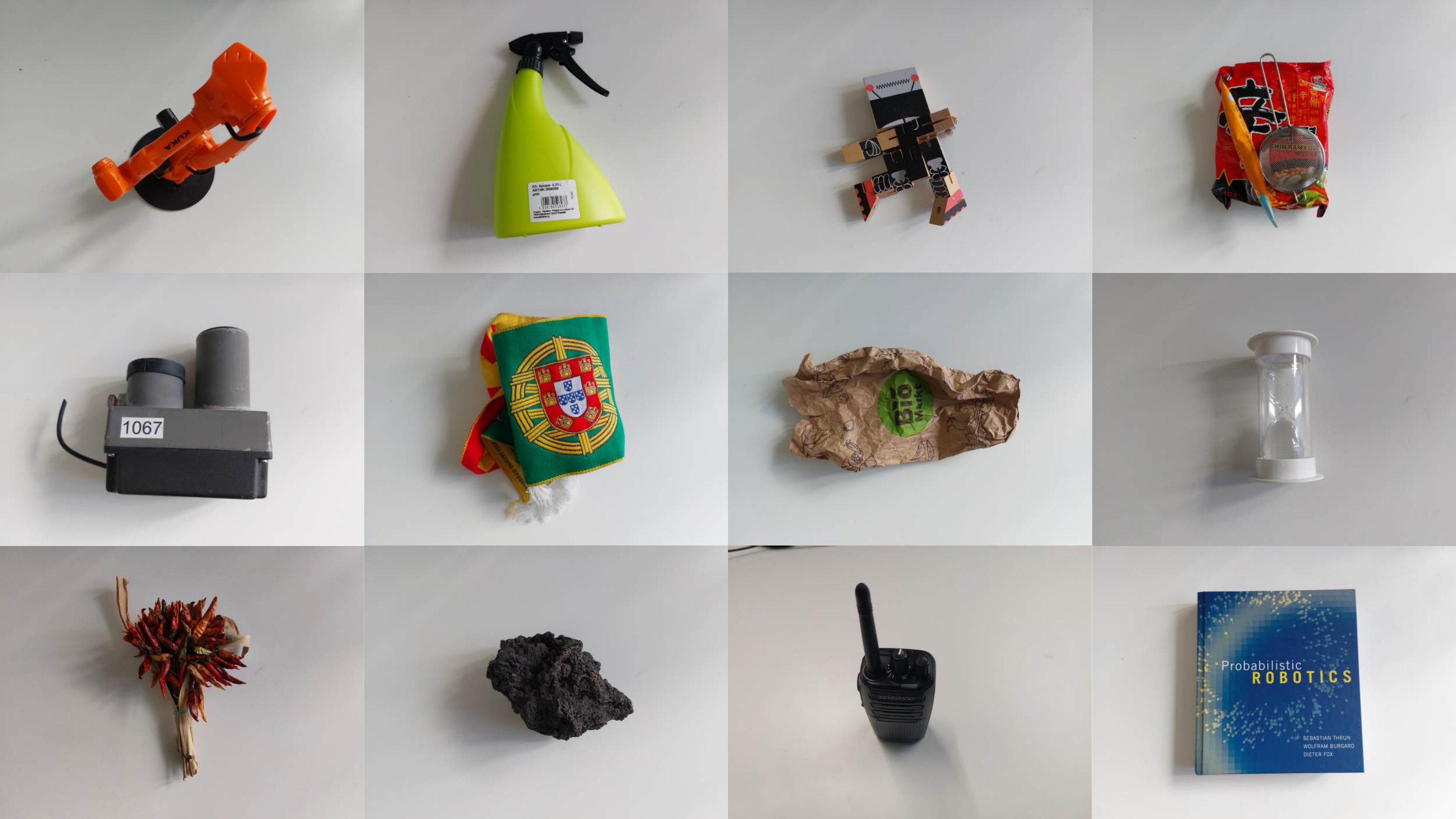}
    \caption{Examples of arbitrary objects brought by different users. The objects ranged from articulated, transparent, deformable, industrial and planetary objects.}
    \label{fig:6}
\end{figure}

\subsection{A complete open-set evaluation with users}
\label{sec:results2}
We evaluate the performance of CLEVER in an open-set condition, where the model encounters unseen objects. Furthermore, the goal is to perform user validation in order to show that many users can use the system successfully. To achieve this goal, we randomly invited 13 users and asked them to bring any object of their choice. Fig. \ref{fig:6} shows the example objects that were brought by the users to test the system. Since we did not know these objects a priori, we could create a truly open-set condition. \blue{Initial prior from Section \ref{sec:results1} was used by pretraining with object and no-object classes.} The users were instructed to use CLEVER in order to perform semantic perception of the objects they brought. Under these conditions, we measured the number of queries to learn the new object with more than 85$\%$ confidence, the number of query failures, test ECE, and training time. The users only collected 80 images per query step. Out of 80 images, CLEVER selected 32 images to adapt the model. \blue{In each query step, posterior of the previous task was used as prior, along with an optimization for PAC-Bayes bounds.}

The results are depicted in Fig. \ref{fig:5}. The average number of required queries was $2.0\pm1.79$, while we had a query success rate of $91.20\pm15.06 \%$ where CLEVER associated well-calibrated confidence and appropriately asked for help. The average ECE was $5.36 \pm 3.75 \%$, and CLEVER consumed $17.79\pm4.717$s training time. All users were able to work with CLEVER, and perform stream-based AL under the replicated open-set condition. We also believe that a training time of less than 20s can be practical. Regarding limitations, small distributional shifts seem to be an issue. For example, we found it difficult to obtain well-calibrated uncertainty estimates when training a DNN with an apple but testing with an object similar to an apple in appearance, like a red pear. Nevertheless, all the objects brought by the user could be eventually conquered with CLEVER, which could have been difficult without adaptations at test-time. In this sense, our experiments show the relevance of stream-based AL in developing a persistent vision system.

\subsection{Demonstration on a humanoid robot}
\label{sec:results3}
Finally, we demonstrate CLEVER on the KIT's humanoid robot, ARMAR-6 \cite{asfour2019armar} (see Fig. \ref{fig1:teaser}) where we examine the feasibility of stream-based AL on a real robot. Three objects, namely an apple, a banana, and a T-shirt, are considered. ARMAR-6's onboard cameras, speech interface, and NVIDIA GeForce GTX 1080 are utilized \cite{asfour2019armar}. Pre-designed language prompts were used to allow robot-human communication. The videos are on our project website. On ARMAR-6, we showcase CLEVER's ability to perform robust semantic perception. We emphasize that, for all examined scenarios, deploying a standard DNN without any ability to adapt would have failed to complete the given perception tasks. In contrast, CLEVER is able to improve the robustness of deploying DNNs on a real robot. That is, CLEVER estimates uncertainties, asks for help from humans, and adapts itself to finally accomplish the given perception task. With these results, we illustrate robust DNN-based perception by showing the feasibility of stream-based AL on a real robot. \footnote{\blue{All implementation details including metrics, synthetic data generation and in-depth discussions are in suplementary materials of our project website.}} 
\section{Conclusion}
In this work, we propose a stream-based active learner for robust semantic perception with robots. Experimentally, ablation studies provide the insights behind the system's design. We also evaluate CLEVER in an open-set condition with a user validation, in which participants brought various objects that are transparent, deformable, articulated, industrial, and planetary. CLEVER is also integrated into a humanoid robot. These results generally suggest the possibilities of robust semantic perception, while embracing the predictive performance of deep learning. In future, improvements in unknown object segmentation techniques will also help our system. Thus, as a next step, we envision active learning on foundational models directly for open-set recognition tasks. 


\bibliographystyle{IEEEtran}
\bibliography{IEEEabrv,bibliography}

\end{document}